\documentclass[runningheads]{llncs}
\usepackage{graphicx}
\usepackage[labelfont=bf]{caption}

\usepackage{siunitx}
\usepackage{amsmath}
\usepackage{amssymb}

\usepackage[caption=false]{subfig}
\usepackage{array}

\usepackage[utf8]{inputenc}
\usepackage{pgfplots}
\usepgfplotslibrary{groupplots,dateplot}
\usetikzlibrary{patterns,shapes.arrows}
\pgfplotsset{compat=newest}

\usepackage{hyperref}
\usepackage{prettyref}

\usepackage[nolist]{acronym}

\usepackage{orcidlink}

\usepackage{multirow}

\usepackage{hyphenat}
\newrefformat{sec}{Section~\ref{#1}}
\newrefformat{tab}{Table~\ref{#1}}
\newrefformat{fig}{Fig.~\ref{#1}}
\newrefformat{subsec}{Subsection~\ref{#1}}
\newrefformat{eq}{(\ref{#1})}

\begin{document}

\begin{acronym}
    \acro{VLAD}{Vector of Locally Aggregated Descriptors}
    \acro{CNN}{Convolutional Neural Network}
    \acro{mAP}{Mean Average Precision}
    \acro{ESVM}{Exemplar-SVM}
    \acro{SGR}{Similarity Graph Reranking}
    \acro{QE}{Query Expansion}
    \acro{kRNN}{$k$ Reciprocal Nearest Neighbor}
\end{acronym}

\title{Towards Writer Retrieval for Historical Datasets}
\titlerunning{Towards Writer Retrieval for Historical Datasets}


\author{Marco Peer\orcidlink{0000-0001-6843-0830}\and
Florian Kleber \orcidlink{0000-0001-8351-5066}\and
Robert Sablatnig\orcidlink{0000-0003-4195-1593} 
}
\authorrunning{M. Peer, F. Kleber and R. Sablatnig}

\institute{Computer Vision Lab\\
TU Wien\\
\email{\{mpeer, kleber, sab\}@cvl.tuwien.ac.at}\\
Code: \url{https://github.com/marco-peer/icdar23}
}

\maketitle        

\begin{abstract}
This paper presents an unsupervised approach for writer retrieval based on clustering SIFT descriptors detected at keypoint locations resulting in pseudo-cluster labels. With those cluster labels, a residual network followed by our proposed NetRVLAD, an encoding layer with reduced complexity compared to NetVLAD, is trained on $32\times 32$ patches at keypoint locations. Additionally, we suggest a graph-based reranking algorithm called SGR to exploit similarities of the page embeddings to boost the retrieval performance. Our approach is evaluated on two historical datasets (Historical-WI and HisIR19). We include an evaluation of different backbones and NetRVLAD. It competes with related work on historical datasets without using explicit encodings. We set a new State-of-the-art on both datasets by applying our reranking scheme and show that our approach achieves comparable performance on a modern dataset as well.

\end{abstract}

\section{Introduction}

Writer retrieval is the task of retrieving documents written by the same author within a dataset by finding similarities in the handwriting \cite{keglevic}. In particular, writer retrieval enables experts in history or paleography to trace individuals or social groups across different time epochs \cite{icdar19}. Furthermore, it helps to identify documents of unknown writers and to detect similarities within those documents \cite{unsupervised_icdar17}. Due to the time-consuming process of analyzing large corpora of documents required by experts, image retrieval algorithms are applied to find all relevant documents of a specific writer.

State-of-the-art methods for writer retrieval consist of four parts: First, characteristics of the handwriting within the document are sampled, e.g., by using interest point detectors such as SIFT \cite{keglevic,peer_netmvlad,rasoulzadeh}. Then, traditional algorithms or deep-learning-based approaches are applied to extract features. In the end, those embeddings are encoded and aggregated to obtain powerful global page descriptors, which are then compared to retrieve a ranked list for each query document. Since the datasets contain a training and a test set with disjunct writers, the performance of writer retrieval approaches is evaluated by using each document of the test set as a query once.

\begin{figure}[t]
    \centering
    \includegraphics[width=\textwidth]{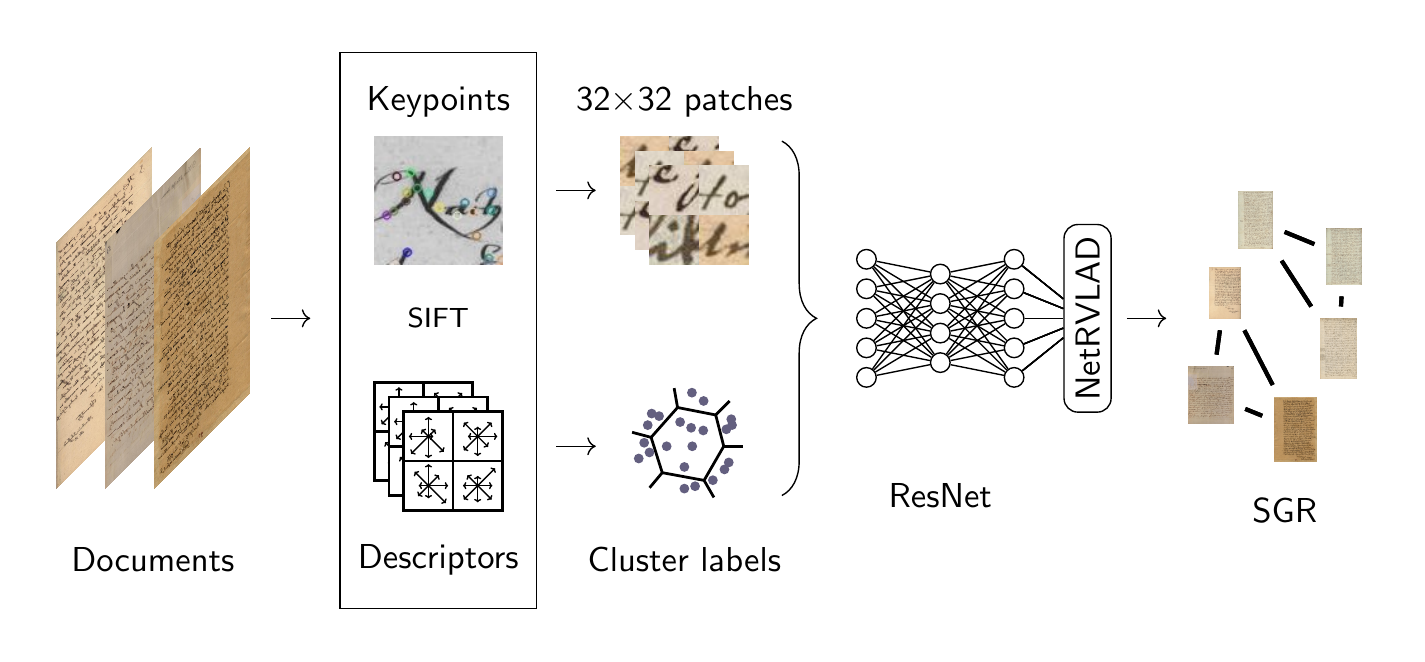}
    \caption{Overview of our proposed pipeline.}
    \label{fig:overview}
\end{figure}

While on modern datasets, neural networks trained in a supervised manner dominate \cite{fiel_cnn,keglevic,peer_netmvlad,rasoulzadeh}, for historical datasets, training on writer label information \cite{peer_selfsupervised,wang_supervised} trails either unsupervised methods \cite{unsupervised_icdar17} or approaches based on handcrafted features \cite{bVLAD}. Historical data introduces additional challenges, e.g., degradation, different languages, the amount of text, or even potential writer-label noise by external influences on handwriting, such as the pen used. However, a different strategy we investigate to improve the performance of writer retrieval is \emph{reranking}: After the global descriptors are calculated and compared, reranking exploits the geometric relationships in the embedding space, as well as the information included in the ranked list to refine the final ranking \cite{reranking_jordan}.

Our paper presents an unsupervised approach illustrated in \prettyref{fig:overview}. It is based on a \ac{CNN} trained on $32\times 32$ patches extracted at SIFT keypoint locations. As a target label, 5000 classes are generated by clustering the corresponding descriptors via k-means \cite{unsupervised_icdar17}. We encode the embeddings of our neural network by Random NetVLAD (NetRVLAD), particularly designed for writer retrieval by removing normalization layers and the initialization, which, we show in our evaluation, harms the performance. In contrast to \cite{unsupervised_icdar17}, we do not rely on external codebooks such as VLAD. Instead, we directly learn a codebook during the network training within the NetRVLAD layer. The global page descriptors are obtained by sum pooling. Secondly, we rerank our global page descriptors with our proposed \ac{SGR} and boost the performance of NetRVLAD. Our reranking is based on the work of Zhang et al. \cite{graphrerank}, who build a graph and aggregate its vertices to refine the features. Their method relies on two hyperparameters  ($k_1$, $k_2$) dependent on the test set whose properties are usually unknown. We propose \ac{SGR}, where an initial graph of the global page descriptors is built using cosine similarity and a weighting function. Afterward, a graph network refines and aggregates the node features, which are then considered as the reranked descriptors. \ac{SGR} improves the work in \cite{graphrerank} by eliminating $k_1$  and is also robust to the choice of $k_2$ across datasets. Additionally, our results show that when using NetRVLAD, the performance is significantly improved by removing complexity compared to the original NetVLAD. In our experiments, NetRVLAD performs stable, even when choosing a smaller codebook size, reducing computational resources. Combined with \ac{SGR}, we outperform related work. Ultimately, we show that our approach is feasible for smaller modern datasets.

Summarizing, our contributions are:
\begin{itemize}
    \item NetRVLAD, an encoding layer for writer retrieval based on NetVLAD \cite{netvlad},
    \item \ac{SGR}, a reranking algorithm using a similarity graph,
    \item a thorough evaluation of our approach on two historical datasets, namely the Historical-WI \cite{icdar17}, and HisIR19 \cite{icdar19}, where we outperform State-of-the-art on both datasets.
\end{itemize}

The remaining part of our paper is structured as follows: In \prettyref{sec:rw}, we describe the related work regarding writer retrieval and reranking strategies used. We cover our approach, including NetRVLAD and \ac{SGR} in \prettyref{sec:method} followed by our evaluation protocol and implementation details in \prettyref{sec:eval}. Our experiments and results are given in \prettyref{sec:exp}. We conclude our paper in \prettyref{sec:conclusion}.

\section{Related Work}\label{sec:rw}
In the following, we give an overview of related work for writer retrieval as well as reranking strategies.

\subsection{Writer Retrieval}

Writer retrieval approaches are divided into codebook-based and codebook-free methods. Those codebooks are used as a model to calculate statistics of the handwriting, with \ac{VLAD} the most prominent one for writer retrieval \cite{zernike,unsupervised_icdar17,christlein_cnn_vlad,keglevic}. Additionally, the characteristics of the handwriting are either extracted by traditional algorithms (handcrafted features) or deep learning. 

For codebook-based methods on modern datasets such as ICDAR2013~\cite{icdar2013} or CVL~\cite{CVL}, the authors of \cite{gmm} compute SURF features encoded by Gaussian mixture models. Christlein et al. \cite{zernike} extract Zernike moments of the contours and build a codebook based on multiple \ac{VLAD} encodings. In contrast to those handcrafted features, Fiel and Sablatnig \cite{fiel_cnn} introduced \acp{CNN} to the domain of writer retrieval. Their codebook-free method relies on aggregating \ac{CNN} activations via sum-pooling. Similarly, \acp{CNN} are applied in \cite{christlein_cnn_vlad,keglevic} as a feature extractor followed by \ac{VLAD}. The authors of \cite{peer_netmvlad,rasoulzadeh} investigate NetVLAD \cite{netvlad}, a learnable version of \ac{VLAD}, plugged in at the end of the network to directly learn the codebook during training. All of those networks are trained in a supervised manner with the writer label as target.

For historical datasets, Christlein et al. \cite{unsupervised_icdar17} show that training on pseudolabels generated by clustering the SIFT descriptors outperforms supervised methods such as \cite{wang_supervised}. Furthermore, for each descriptor, an \ac{ESVM} is trained to refine the encoding. Peer et al. \cite{peer_selfsupervised} apply a self-supervised algorithm using morphological operations to generate augmented views without any labels. The winners of the HisIR19 competition \cite{icdar19} and the current state-of-the-art method on the Historical-WI dataset rely on handcrafted features (SIFT and pathlet) for retrieval. They encode both features via \emph{bagged VLAD} (bVLAD) \cite{bVLAD}. Our approach is mainly inspired by the work in \cite{unsupervised_icdar17}, but we train our network with triplets and additionally encode our embeddings based on NetVLAD.

\subsection{Reranking}
While reranking is a method to improve the performance of image retrieval in general, two approaches \cite{reranking_jordan,rasoulzadeh} investigate reranking in the domain of writer retrieval. 

In \cite{rasoulzadeh}, Rasoulzadeh and Babaali propose an adaption to the standard reranking method \ac{QE} \cite{QE}. They average each descriptor with their top \ac{kRNN} and show that they can boost the retrieval performance by reducing the effect of false matches. Jordan et al. \cite{reranking_jordan} extend the \acp{ESVM} of Christlein et al. \cite{unsupervised_icdar17} as a baseline for their reranking evaluation. They consider additional positive samples for the training \acp{ESVM} called \emph{Pair} or \emph{Triple SVM} and increase the performance of \cite{unsupervised_icdar17}. 

Recent methods in image retrieval apply neural networks to refine the ranking, e.g., Tan et al. \cite{reranking_transformers} suggest reranking transformers, and Gordo et al. propose attention-based query expansion learning with a contrastive loss \cite{attention_reranking}.  

Our approach is based on the work of Zhang et al. \cite{graphrerank}. They build a graph with the $k_1$ nearest neighbors and aggregate the nodes of the $k_2$ nearest neighbors by using a graph network, arguing the generality of their approach, e.g., including approaches like $\alpha$-\ac{QE} \cite{alphaQE}. We suggest using the similarity of the embeddings to create the initial graph, which removes the requirement for selecting an appropriate value for $k_1$.

\section{Methodology}\label{sec:method}

In this section, we describe each aspect of our approach and explain the two main parts we propose for writer retrieval: NetRVLAD and \ac{SGR}.

\subsection{Patch Extraction}

\begin{figure}[t]
    \centering
    \subfloat{
        \centering
        \includegraphics[width=0.05\textwidth]{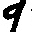}
        \includegraphics[width=0.05\textwidth]{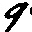}
        \includegraphics[width=0.05\textwidth]{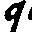}
        \includegraphics[width=0.05\textwidth]{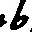}
        \includegraphics[width=0.05\textwidth]{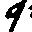}
        \includegraphics[width=0.05\textwidth]{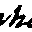}
        \includegraphics[width=0.05\textwidth]{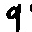}
        \includegraphics[width=0.05\textwidth]{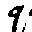}
    }
    \subfloat{ 
        \centering
        \includegraphics[width=0.05\textwidth]{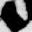}
        \includegraphics[width=0.05\textwidth]{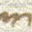}
        \includegraphics[width=0.05\textwidth]{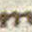}
        \includegraphics[width=0.05\textwidth]{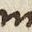}
        \includegraphics[width=0.05\textwidth]{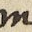}
        \includegraphics[width=0.05\textwidth]{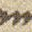}
        \includegraphics[width=0.05\textwidth]{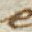}
        \includegraphics[width=0.05\textwidth]{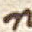}

}

\hspace{0.1cm}
\setcounter{subfigure}{0}

    \subfloat[Historical-WI]{
        \includegraphics[width=0.05\textwidth]{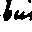}
        \includegraphics[width=0.05\textwidth]{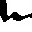}
        \includegraphics[width=0.05\textwidth]{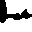}
        \includegraphics[width=0.05\textwidth]{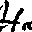}
        \includegraphics[width=0.05\textwidth]{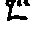}
        \includegraphics[width=0.05\textwidth]{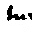}
        \includegraphics[width=0.05\textwidth]{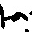}
        \includegraphics[width=0.05\textwidth]{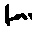}
    }
    \subfloat[HisIR19]{
        \includegraphics[width=0.05\textwidth]{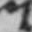}
        \includegraphics[width=0.05\textwidth]{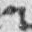}
        \includegraphics[width=0.05\textwidth]{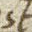}
        \includegraphics[width=0.05\textwidth]{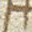}
        \includegraphics[width=0.05\textwidth]{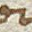}
        \includegraphics[width=0.05\textwidth]{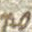}
        \includegraphics[width=0.05\textwidth]{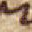}
        \includegraphics[width=0.05\textwidth]{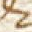}
    }
    
    \caption{Examples of the clustered $32\times32$ patches.}\label{fig:patch_examples}    
\end{figure}

Our preprocessing is based on the approach of Christlein et al. \cite{unsupervised_icdar17}. Firstly, we detect keypoints for each document as well as the corresponding descriptors, both via SIFT. These descriptors are normalized with the Hellinger kernel (elementwise square root followed by $l_1$-normalization) and dimensionality reduction via PCA from 128 to 32. We cluster the descriptors via k-means in 5000 clusters \cite{unsupervised_icdar17}. As an additional preprocessing step, we filter keypoints whose descriptors $\boldsymbol{\mathrm{d}}$ violate

\begin{equation}\label{eq:filter}
    \frac{|| \boldsymbol{\mathrm{d}} - \boldsymbol{\mathrm{\mu}}_1||}{||\boldsymbol{\mathrm{d}} - \boldsymbol{\mathrm{\mu}}_2 ||} > \rho,
\end{equation}

where $\boldsymbol{\mathrm{\mu}}_i$ denotes the $i$-th nearest cluster of $\boldsymbol{\mathrm{d}}$ and $\rho = 0.9$. By applying \prettyref{eq:filter} we filter keypoints that lay near the border of two different clusters - those are therefore considered to be ambiguous. The $32\times32$ patch is extracted at the keypoint location, and the cluster membership is used as a label to train the neural network. In \prettyref{fig:patch_examples}, we show eight samples of two clusters each for both datasets used. We observe clusters where characters written in a specific style dominate, e.g., 'q' or 'm' on top, and clusters containing general patterns included in the handwriting (bottom).

\subsection{Network architecture}

Our network consists of two parts: a residual backbone and an encoding layer, for which we propose \emph{NetRVLAD}. The output of NetRVLAD is used as a global descriptor of the $32\times 32$ input patch.

\paragraph{Residual backbone} Similar to  \cite{unsupervised_icdar17,rasoulzadeh}, the first stage of our network is a ResNet to extract an embedding for each patch. The last fully connected layer of the network is dropped, and the output of the global averaging pooling layer of dimension $(64,1,1)$ is used. We evaluate the choice of the depth of the network in our results.

\paragraph{NetRVLAD} The traditional \ac{VLAD} algorithm clusters a vocabulary to obtain $N_c$ clusters $\{ \boldsymbol{\mathrm{c}}_0, \boldsymbol{\mathrm{c}}_1, \dots, \boldsymbol{\mathrm{c}}_{N_c-1}\}$ and encodes a set of local descriptors $\boldsymbol{\mathrm{x}}_i,$ $i \in \{0, \dots, N-1\}$ via

\begin{equation}\label{eq:vlad}
	\boldsymbol{\mathrm{v}}_k =  \sum_{i=0}^{N-1} \boldsymbol{\mathrm{v}}_{k,i} = \sum_{i=0}^{N-1} {\alpha}_k(\boldsymbol{\mathrm{x}}_i)  (\boldsymbol{\mathrm{x}}_i -  \boldsymbol{\mathrm{c}}_k), \quad k \in \{0, \dots, N_c-1\},
\end{equation}

with $\alpha_k = 1$  if $\boldsymbol{\mathrm{c}}_k$ is the nearest cluster center to $\boldsymbol{\mathrm{x}}_i$, otherwise 0, hence making the VLAD encoding not differentiable. The final global descriptor is then obtained by concatenating the vectors $\boldsymbol{\mathrm{v}}_k$. Arandjelović et al. \cite{netvlad} suggest the NetVLAD layer which tackles the non-differentiability of $\alpha_k$ in \prettyref{eq:vlad} by introducing a convolutional layer with parameters $\{\boldsymbol{\mathrm{w}}_{k},b_{k}\}$ for each cluster center $\boldsymbol{\mathrm{c}}_k$ to learn a soft-assignment

\begin{equation}
	\overline{\alpha}_k(\boldsymbol{\mathrm{x}}_i) = \frac{e^{\boldsymbol{\mathrm{w}}_k^\mathrm{T} \boldsymbol{\mathrm{x}}_i + b_k }}{\sum_{k'}^{} e^{\boldsymbol{\mathrm{w}}_{k'}^\mathrm{T} \boldsymbol{\mathrm{x}}_i + b_{k'} }}.
\end{equation}

\begin{figure}[t] \centering
    \subfloat[Traditional \ac{VLAD}]{
        \includegraphics[width=0.48\textwidth]{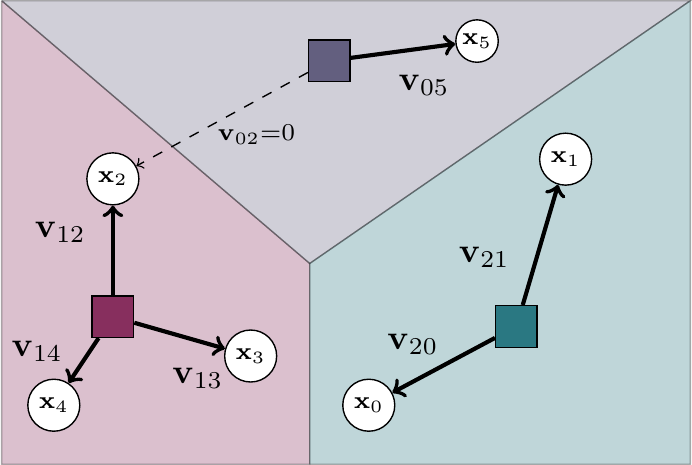}
    }
    \subfloat[NetVLAD]{
        \includegraphics[width=0.48\textwidth]{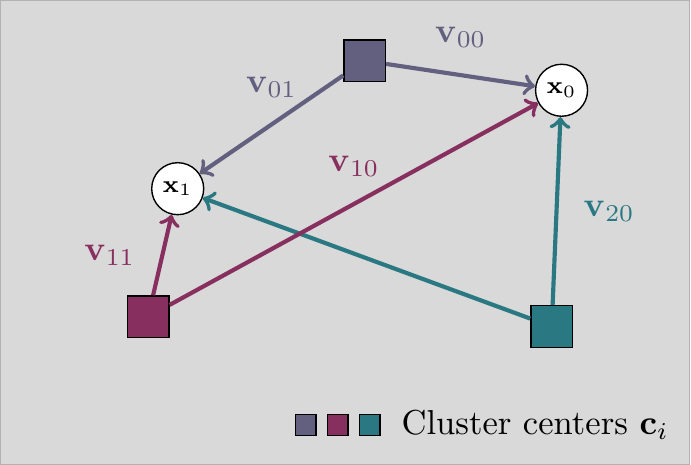}
    }
    \caption{Traditional \ac{VLAD} and NetVLAD. While \ac{VLAD} hard-assigns each descriptor $x_i$ to its nearest cluster to compute the residual, NetVLAD directly learns 1) the cluster centers and 2) their assignments allowing to aggregate multiple residuals for one descriptor.} \label{fig:vlad}
\end{figure}

The cluster centers $\boldsymbol{\mathrm{c}}_k$ are also learned during training. A schematic comparison is shown in \prettyref{fig:vlad}. The input of NetVLAD is a feature map of dimension $(D, H, W)$ handled as a $D\times N$ spatial descriptor with $N=HW$. Normalization and concatenation of the vectors $\boldsymbol{\mathrm{v}}_k$  

\begin{align}
    \boldsymbol{\mathrm{v}}_{k} = \sum_{i=0}^{N} \boldsymbol{\mathrm{v}}_{k,i} = \sum_{i=0}^{N} \overline{\alpha}_k(\boldsymbol{\mathrm{x}}_i)  (\boldsymbol{\mathrm{x}}_i -  \boldsymbol{\mathrm{c}}_k), \quad k \in \{0, \dots, N_c-1\}
\end{align}
yields the final NetVLAD encoding $\boldsymbol{\mathrm{V}} \in \mathbb{R}^{N_c \times D}$.

For writer retrieval, the main idea of applying NetVLAD is learning a powerful codebook via its cluster centers, representing, e.g., features like characters or combinations of them or more high-level ones like slant directions of the handwriting. We generate a meaningful descriptor by concatenating the residuals between a patch embedding to the cluster centers. A page is then characterized by measuring differences between those features. In contrast to VLAD, the codebook is directly integrated into the network. For our approach, we reduce the complexity of NetVLAD and adapt two aspects which we call \emph{NetRVLAD}: 

1) Similar to RandomVLAD proposed by Weng et al. \cite{randomvlad}, we loosen the restriction of the embeddings $\boldsymbol{\mathrm{x}}$ of the backbone as well as the cluster residuals $\boldsymbol{\mathrm{v}}_{k}$ lying on a hypersphere. Since we only forward one descriptor per patch ($H=W=1$), we argue that NetVLAD learns this during training on its own - therefore, we remove the pre- and intranormalization of NetVLAD. 

2) Arandjelovic et al. \cite{netvlad} propose an initialization of the convolutional layer where the ratio of the two closest (maximum resp. second highest value of $\overline{\alpha}_k$) cluster assignments is equal to ${\alpha}_\mathrm{init} \approx 100$. To improve performance, we initialize the weights of the convolutional layer and the cluster centers randomly rather than using a specific initialization method, as this can increase the impact of the initialization of the cluster centers. Additionally, the hyperparameter ${\alpha}_\mathrm{init}$ is removed. We compare NetRVLAD to the original implementation in \prettyref{sec:eval}.


\subsection{Training} Our network is trained with the labels assigned while clustering the SIFT descriptors. Each patch is embedded in a flattened $N_c \times 64$ descriptor. We directly train the encoding space using the distance-based triplet loss

\begin{equation}\label{eq:triplet}
	\mathcal{L}_\text{Triplet} = \text{max}(0, d_{ap} - d_{an} + m),
\end{equation}

with the margin $m$ where $a$ denotes the anchor, $p$ the positive and $n$ the negative sample. We only mine \emph{hard} triplets \cite{triplets} in each minibatch. Therefore, each triplet meets the criterion.

\begin{equation}
    d_{an} < d_{ap} - m.
\end{equation}

\subsection{Global Page Descriptor}

During inference, we aggregate all embeddings $\{\boldsymbol{\mathrm{V}}_0, \boldsymbol{\mathrm{V}}_1, \dots, \boldsymbol{\mathrm{V}}_{n_p-1}\}$ of a page using $l_2$ normalization followed by sum pooling

\begin{equation}
    \boldsymbol{\mathrm{V}} = \sum_{i=0}^{n_p-1} \boldsymbol{\mathrm{V}}_i
\end{equation}

to obtain the global page descriptor $\boldsymbol{\mathrm{V}}$. Furthermore, to reduce visual burstiness \cite{visual_burstiness}, we apply power-normalization $f(x) = \text{sign}(x)|x|^\alpha$ with $\alpha = 0.4$, followed by $l_2$-normalization. Finally, a dimensionality reduction with whitening via PCA is performed. 


\subsection{Reranking with \ac{SGR}}
Writer retrieval is evaluated by a leave-one-out strategy: Each image of the set is once used as a query $q$, the remaining documents are called the gallery. For each $q$, the retrieval returns a ranked list of documents $\mathrm{L}(q)$. Reranking strategies exploit the knowledge contained in $\mathrm{L}(p_i)$ with $ p_i \in \mathrm{L}(q)$ and refine the descriptors \cite{reranking_jordan}. We can intuitively model those relationships by a graph $\mathcal{G} = \big(\mathcal{V}, \mathcal{E}\big)$ with its vertices $\mathcal{V}$ and edges $\mathcal{E}$.

Our approach called \ac{SGR} is conceptually simple and consists of two stages inspired by the work in \cite{graphrerank}.  The first stage is building the initial graph using the page descriptors to compute the vertices. Instead of only considering $k$ nearest neighbours as described by Zhang et al. \cite{graphrerank}, we propose to use the cosine similarity $s_{i,j} = \boldsymbol{\mathrm{x}}^\mathrm{T}_i \cdot  \boldsymbol{\mathrm{x}}_j$ and obtain the symmetric adjacency matrix by
\begin{equation}\label{eq:sim}
    \boldsymbol{\mathrm{A}}_{i,j} = \mathrm{exp}\big(-\frac{(1-s_{i,j})^2}{\gamma}\big)
\end{equation}
with a hyperparameter $\gamma$ which mainly determines the decay of edge weights when similarity decreases. Therefore, our approach additionally benefits from a continuous adjacency matrix by using the learned embedding space. We consider similarities while replacing the task-dependent hyperparameter $k_1$ in \cite{graphrerank}.


Furthermore, we compute the vertices by encoding the similarity of each descriptor instead of adopting the original page descriptors: The rows of the adjacency matrix $\boldsymbol{\mathrm{A}}$ - we denote the $i$-th row as $\boldsymbol{\mathrm{h}}_i$ in the following - are used as page descriptors which we refer to as a \emph{similarity graph}. While Zhang et al. \cite{graphrerank} propose a discrete reranked embedding space ($\boldsymbol{\mathrm{A}}_{i,j} \in \{0, \frac{1}{2}, 1\}$), we argue that a continuous embedding space further improves the reranking process by using our weighting function to refine the embeddings. Thus, we are able to exploit not only the neighborhood of a page descriptor, but also its distances.

Secondly, each vertex is propagated through a graph network consisting of $L$ layers via

\begin{equation}
    \boldsymbol{\mathrm{h}}_i^{(l+1)} = \boldsymbol{\mathrm{h}}_i^{(l)} + \sum_j s_{i,j} \ \boldsymbol{\mathrm{h}}_j^{(l)}, \quad j \in \mathcal{N}(i, k), \quad l \in \{1, \dots, L\},
\end{equation}

 where $\mathcal{N}(i, k)$ denotes the $k$ nearest neighbors of vertex $i$. Those neighbors are aggregated with their initial similarity $s_{i,j}$. During message propagation, we only consider the $k$ (equal to $k_2$ in \cite{graphrerank}) nearest neighbors to reduce the noise of aggregating wrong matches ($k$ is usually small, e.g., $k=\num{2}$) and also eliminate the influence of small weight values introduced by \prettyref{eq:sim}. The vertices are $l_2$-normalized after each layer. $\boldsymbol{\mathrm{h}}_i^{(L)}$ is used as the final reranked page descriptor. In our evaluation, we report the performances of SGR, as well as the initial approach by Zhang et al. \cite{graphrerank}, and provide a study on the hyperparameters of our reranking.


\section{Evaluation Protocol}\label{sec:eval}
In this section, we cover the datasets and metrics used and give details about our implementation.


\subsection{Datasets}
\begin{figure}
    \centering
    \subfloat[Historical-WI]{
        \centering
        \includegraphics[width=0.25\textwidth]{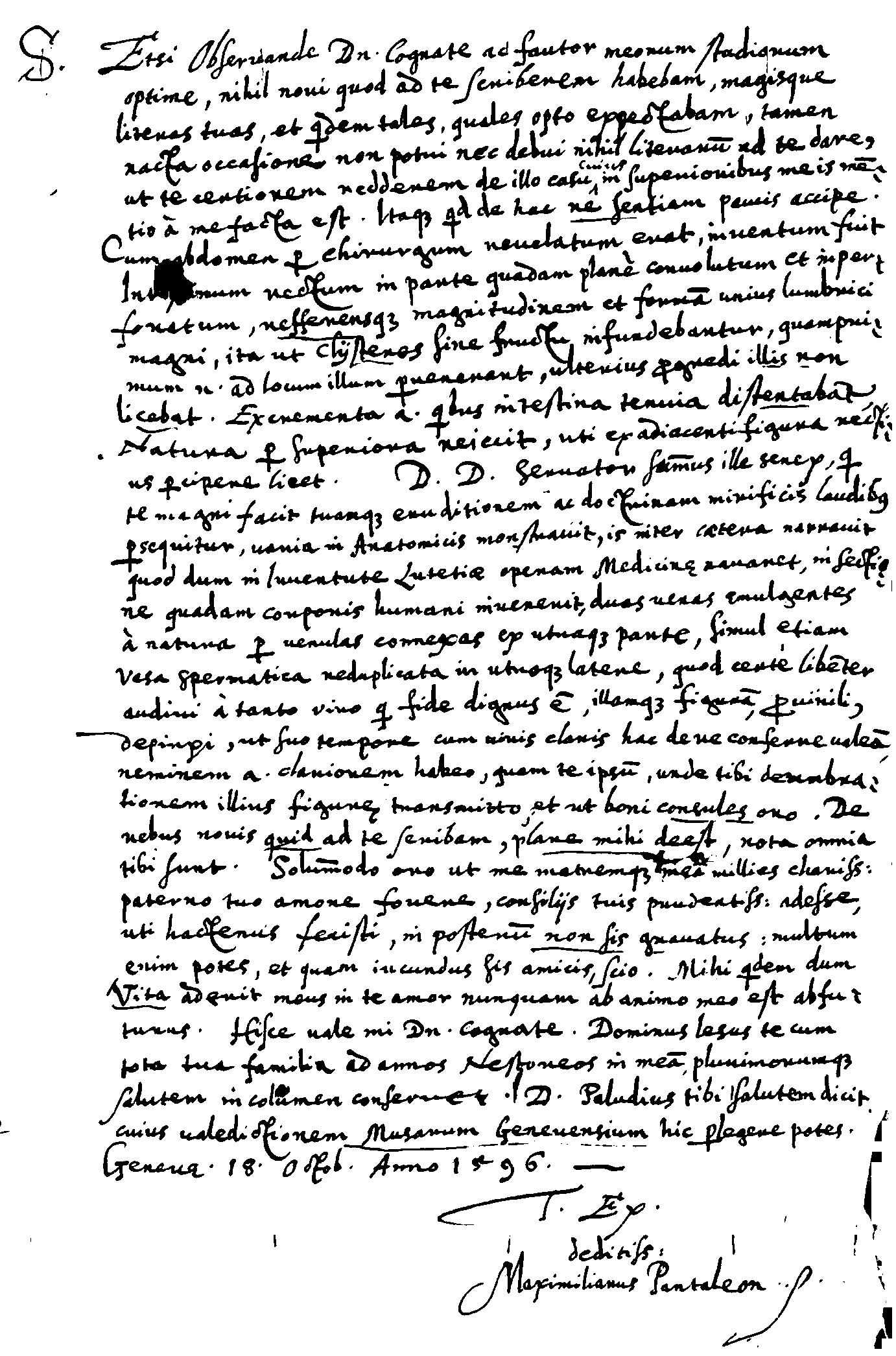}
    }
    \hfill
    \subfloat[HisIR19]{
        \centering
        \includegraphics[width=0.3\textwidth]{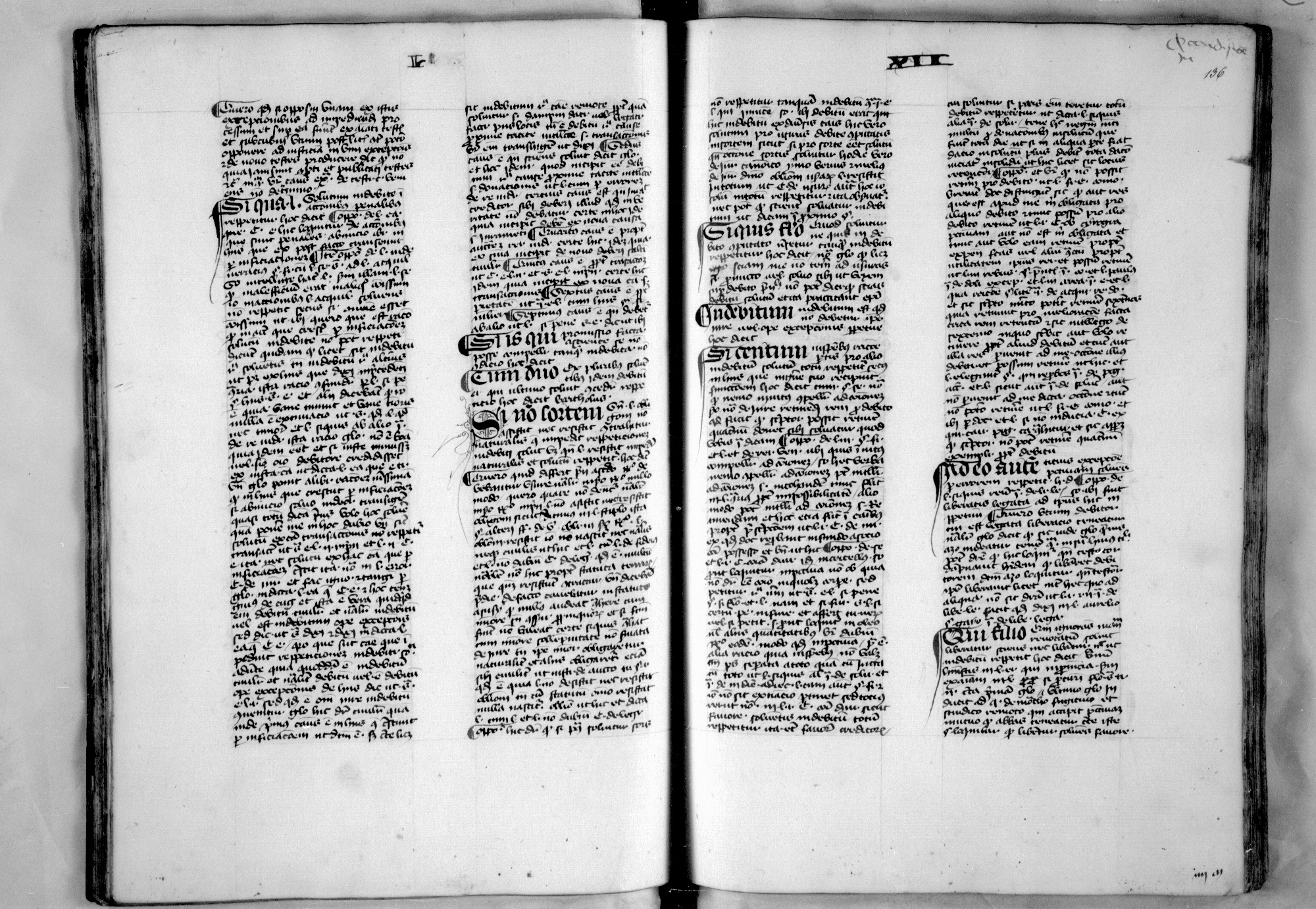}
        
        \includegraphics[width=0.3\textwidth]{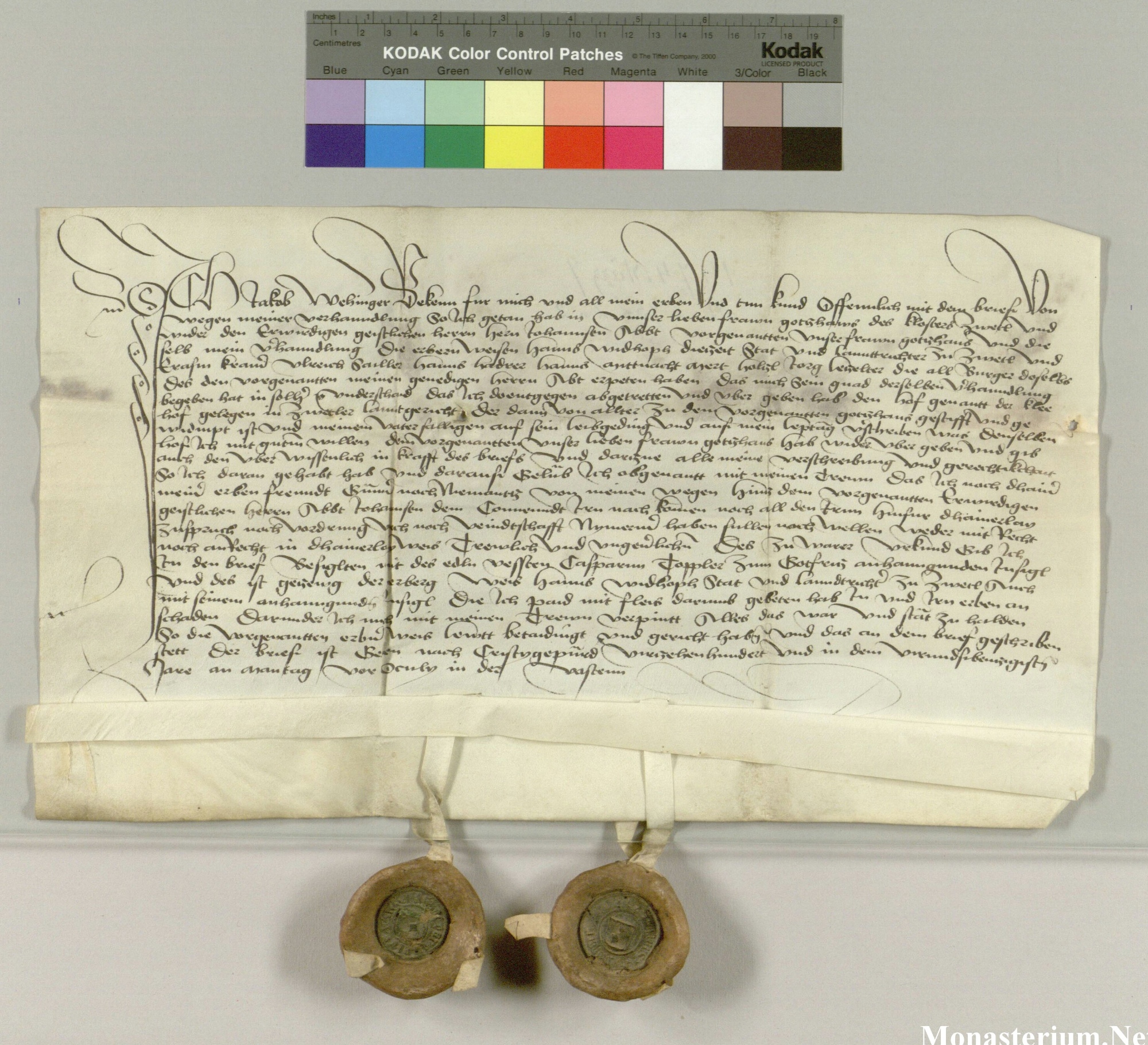}
    }
    
    \caption{Example images of the datasets used.}\label{fig:dataset_images}    
\end{figure}

We use two historical datasets with their details stated in the following. In \prettyref{fig:dataset_images}, examples of the two datasets used are shown.

\paragraph{Historical-WI} This dataset proposed by Fiel et al. \cite{icdar17} at the \emph{ICDAR 2017 Competition on Historical Document Writer Identification} consists of \num{720} authors where each one contributed five pages, resulting in a total of \num{3600} pages. Originating from the 13th to 20th century, the dataset contains multiple languages such as German, Latin, and French. The training set includes \num{1182} document images written by \num{394} writers with an equal distribution of three pages per writer. Both sets are available as binarized and color images. To ensure a fair comparison, we follow related work and report our results on the binarized version of the dataset.
\paragraph{HisIR19} Introduced at the \emph{ICDAR 2019 Competition on Image Retrieval for Historical Handwritten Documents} by Christlein et al. \cite{icdar19}, the test set consists of \num{20000} documents of different sources (books, letters, charters, and legal documents). \num{7500} pages are isolated (one page per author), and the remaining authors contributed either three or five pages. The training set recommended by the authors of \cite{icdar19} and used in this paper is the validation set of the competition, including 1200 images of 520 authors. The images are available in color.

\subsection{Metrics}
To evaluate performance, we use a leave-one-out retrieval method where each document is used as a query and a ranked list of the remaining documents is returned. The similarity is measured by using cosine distance between global page descriptors. 

Our results are reported on two metrics. \ac{mAP} and Top-1 accuracy. While \ac{mAP} considers the complete ranked list by calculating the mean of the average precisions, Top-1 accuracy measures if the same author writes the nearest document within the set. 

\subsection{Implementation Details}

\paragraph{Patch extraction and label generation} For preprocessing, we rely on the experiments of Christlein et al. \cite{unsupervised_icdar17} and use $32\times 32$ patches clustered into 5000 classes. We only use patches with more than $5\%$ black pixels for binary images. To filter the patches of color images, a canny edge detector is applied, and only patches with more than $10 \%$ edge pixels are taken \cite{wang_supervised}. This value is chosen since the HisIR19 dataset contains multiple sources of noise (book covers, degradation of the page, or color palettes) we consider irrelevant for writer retrieval. To decrease the total number of patches of the test sets, we limit the number of patches on a single page to 2000. 

\paragraph{Training} We train each network for a maximum of \num{30} epochs with a batch size of \num{1024}, a learning rate of $l_r = 10^{-4}$ and a margin $m=\num{0.1}$ for the triplet loss. Each batch contains 16 patches per class.  $10\%$ of the training set are used as the validation set. We stop training if the \ac{mAP} on the validation set does not increase for five epochs. Optimization is done with Adam and five warmup epochs during which the learning rate is linearly increased from $l_r / 10$ to $l_r$. Afterward, a cosine annealing is applied. As data augmentation, we apply erosion and dilation. All of our results on the trained networks are averages of three runs with the same hyperparameters but different seeds to reduce the effect of outliers due to initialization or validation split. If not stated otherwise, our default network is ResNet56 with $N_c = 100$.

\paragraph{Retrieval and Reranking} For aggregation, the global page descriptor is projected into a lower dimensional space (performance peaks at \num{512} for Historical-WI, \num{1024} for HisIR19) via a PCA with whitening followed power-normalization ($\alpha=\num{0.4}$) and a $l_2$-normalization. For experiments in which the embedding dimension is smaller than \num{512}, only whitening is applied. 


\section{Experiments}\label{sec:exp}

We evaluate each part of our approach in this section separately, starting with NetRVLAD and its settings, followed by a thorough study of \ac{SGR}.
In the end, we compare our results to state-of-the-art methods on both datasets.
\subsection{NetRVLAD}
Firstly, we evaluate the backbone of our approach. We choose four residual networks of different depth, starting with ResNet20 as in related work \cite{unsupervised_icdar17,rasoulzadeh} up to ResNet110, and compare the performance of NetVLAD to our proposed NetRVLAD. As shown in \prettyref{tab:netrvlad}, NetRVLAD consistently outperforms the original NetVLAD implementation in all experiments. Secondly, we observe deeper networks to achieve higher performances, although, on the Historical-WI dataset, the gain saturates for ResNet110. ResNet56 with our NetRVLAD layer is used for further experiments as a tradeoff architecture between performance and computational resources.

\begin{table}%
    \centering
    \renewcommand{\arraystretch}{1.3}
    \caption{Comparison of NetVLAD and NetRVLAD on different ResNet architectures with $N_c = \num{100}$. Each result is an average of three runs with different seeds.} \label{tab:netrvlad}
  \begin{tabular}{p{0.17\textwidth}cccccccc} \hline
      ~   & \multicolumn{4}{c}{Historical-WI} & \multicolumn{4}{c}{HisIR19}  \\ \cline{2-5} \cline{6-9}
      ~ & \multicolumn{2}{c}{NetRVLAD} & \multicolumn{2}{c}{NetVLAD} &\multicolumn{2}{c}{NetRVLAD} & \multicolumn{2}{c}{NetVLAD} \\ 
      ~ & mAP & Top-1 & mAP & Top-1 &  mAP & Top-1 & mAP & Top-1 \\ \hline
      ResNet20  &  \num{71.5} & \num{87.6} &  \num{67.4} & \num{85.3} & \num{90.1} & \num{95.4} & \num{89.4} & \num{94.5} \\ 
      ResNet32 &  \num{72.1} & \num{88.2} &  \num{67.9} & \num{85.5} & \num{90.6} & \num{95.7}  & \num{89.6} & \num{94.9} \\
      ResNet56 &  $\boldsymbol{73.1}$ & $\boldsymbol{88.3}$ &  \num{68.3} & \num{85.8} & \num{91.2} & \num{96.0} & \num{90.2} &  \num{95.3}  \\ 
      ResNet110 &  $\boldsymbol{73.1}$ & $\boldsymbol{88.3}$  & \num{68.9}  & \num{86.2} &  $\boldsymbol{91.6}$ & $\boldsymbol{96.1}$ &\num{89.9} &  \num{95.5} \\ \hline
    \end{tabular}
\end{table}

\begin{figure}
    \centering
    \subfloat[Historical-WI]{
        \centering
        \begin{tikzpicture}

\definecolor{darkgray176}{RGB}{176,176,176}
\definecolor{steelblue31119180}{RGB}{31,119,180}

\begin{axis}[
width=0.48\textwidth,
tick align=inside,
tick pos=left,
xmode=log,
log basis x={2},
xmin=0, xmax=1200,
xtick={8,16,32,64,128,256,512,1024},
xticklabels={8,16,32,64,128,256,512,1024},
xtick style={color=black},
ymin=69, ymax=74,
ytick style={color=black},
xlabel=Number of clusters $N_c$,
ylabel=\ac{mAP},
xmajorgrids,
ymajorgrids
]
\addplot [semithick, color=steelblue31119180, mark=*]
table {%
8 69.6
16 71.6
32 72.7
64 73.2
100 73.1
128 73.1
256 73.1
512 72.6
1024 72.6
};
\end{axis}

\end{tikzpicture}
    }
    \subfloat[HisIR19]{
        \centering
        \begin{tikzpicture}

\definecolor{darkgray176}{RGB}{176,176,176}
\definecolor{steelblue31119180}{RGB}{31,119,180}

\begin{axis}[
width=0.48\textwidth,
tick align=inside,
tick pos=left,
xmode=log,
log basis x={2},
xmin=0, xmax=1200,
xtick={8,16,32,64,128,256,512,1024},
xticklabels={8,16,32,64,128,256,512,1024},
xtick style={color=black},
ymin=89, ymax=92,
ytick style={color=black},
xlabel=Number of clusters $N_c$,
ylabel=\ac{mAP},
xmajorgrids,
ymajorgrids
]
\addplot [semithick, color=steelblue31119180, mark=*]
table {%
8 89.7
16 90.7
32 91.2
64 91.3
100 91.3
128 91.3
256 91.6
512 91.1
1024 91.4
};

\end{axis}
\end{tikzpicture}
    }
    \caption{Influence of $N_c$ on the performance in terms of \ac{mAP} on the Historical-WI and HisIR19 test dataset.}\label{fig:eval_NC}    
\end{figure}
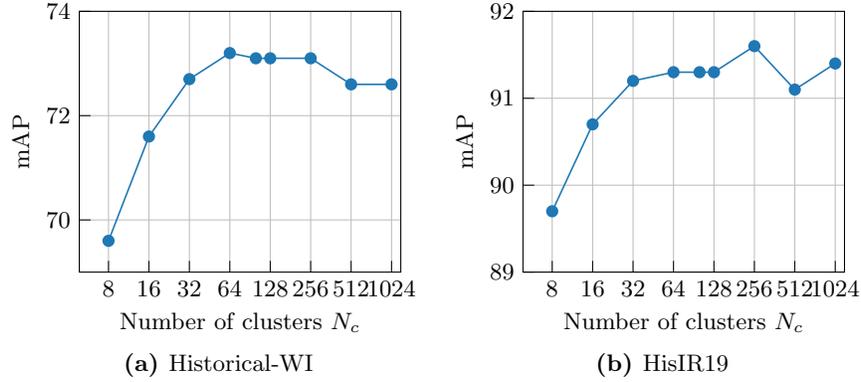

\paragraph{Cluster centers of NetRVLAD} We study the influence of the size $N_c$ of the codebook learned during training. In related work \cite{peer_netmvlad,rasoulzadeh}, the vocabulary size is estimated considering the total amount of writers included in the training set. However, this does not apply to our unsupervised approach. In \prettyref{fig:eval_NC}, we report the performance of NetRVLAD while varying $N_c$. We report a maximum in terms of \ac{mAP} when using a codebook size of 128 resp. 256 on Historical-WI and HisIR19. In general, a smaller codebook works better on Historical-WI; we think this is caused by a) HisIR19 is a larger dataset and b) it introduces additional content, e.g. book covers or color palettes as shown in \prettyref{fig:dataset_images} enabling a better encoding by learning more visual words. For HisIR19, performance is relatively stable over the range we evaluate. Since it also contains noise like degradation or parts of book covers, NetRVLAD seems to benefit when training with more cluster centers. It is also robust - with a small codebook ($N_c = 8$), the drop is only -3.6\% resp. -1.9\% compared to the peak performance. 

\subsection{Reranking}

Once the global descriptors are extracted, we apply \ac{SGR} to improve the performance by exploiting relations in the embedding space by building our similarity graph and aggregating its vertices. \ac{SGR} relies on three hyperparameters: the $k$ nearest neighbors which are aggregated, the number of layers $L$ of the graph network, and $\gamma$, the similarity decay of the edge weights.  While $L$ and $\gamma$ are parameters of the general approach and are validated on the corresponding training set, $k$ is dependent on two aspects: 
\begin{enumerate}
\item The performance of the retrieval on the baseline descriptors - if the top-ranked samples are false, the relevant information within the ranked list is either noise or not considered during reranking. 
\item The gallery size $n_G$ - the number of samples written by an author, either a constant or varies within the dataset.
\end{enumerate}

We evaluate \ac{SGR} by first validating $L$ and $\gamma$ and then studying the influence of $k$ on the test set.

\begin{figure}
    \centering
    \subfloat[Historical-WI]{
        \centering
        \begin{tikzpicture}

\definecolor{darkgray176}{RGB}{176,176,176}
\definecolor{steelblue31119180}{RGB}{31,119,180}

\begin{axis}[
legend style={nodes={scale=0.75, transform shape}},
width=0.48\textwidth,
tick align=inside,
tick pos=left,
xmin=0, xmax=1,
xtick style={color=black},
ymin=65, ymax=85,
ytick style={color=black},
xlabel= $\gamma$,
ylabel=\ac{mAP},
xmajorgrids,
ymajorgrids,
legend style={at={(0.1,0.1)},anchor=south west}
]

\addplot[mark=none, thick, black, samples=2] {79.6};
\addlegendentry{Baseline}
\addplot [semithick, color=steelblue31119180, mark=*]
table {%
0.1 81.8
0.2 82.1
0.3 82.6
0.4 82.7
0.5 82.5
0.6 81.9
0.7 81
0.8 79.2
0.9 73.8
1 63.6
};
\addlegendentry{$L=1$}

\addplot [semithick, color=orange, mark=*]
table {%
0.1 81.8
0.2 82.1
0.3 82.5
0.4 82.5
0.5 81.9
0.6 80.3
0.7 74.3
0.8 66.5
0.9 52.5
1 41.4
};
\addlegendentry{$L=2$}

\addplot [semithick, color=red, mark=*]
table {%
0.1 81.8
0.2 81.9
0.3 82
0.4 81.6
0.5 80.5
0.6 76.5
0.7 67.4
0.8 56.3
0.9 43.7
1 35.6
};
\addlegendentry{$L=3$}

\end{axis}
\end{tikzpicture}
    }
    \subfloat[HisIR19]{
        \centering
        \begin{tikzpicture}

\definecolor{darkgray176}{RGB}{176,176,176}
\definecolor{steelblue31119180}{RGB}{31,119,180}

\begin{axis}[
legend style={nodes={scale=0.75, transform shape}},
width=0.48\textwidth,
tick align=inside,
tick pos=left,
xmin=0, xmax=1,
xtick style={color=black},
ymin=70, ymax=90,
ytick style={color=black},
ytick={70, 75, 80, 85, 90},
yticklabels={70, 75, 80, 85, 90},
xlabel= $\gamma$,
ylabel=\ac{mAP},
xmajorgrids,
ymajorgrids,
legend style={at={(0.1,0.1)},anchor=south west}
]

\addplot[mark=none, thick, black, samples=2] {84.7};
\addlegendentry{Baseline}
\addplot [semithick, color=steelblue31119180, mark=*]
table {%
0.1 86.4
0.2 86.3
0.3 86.3
0.4 86.6
0.5 85.7
0.6 84.3
0.7 81.4
0.8 72.1
0.9 57.7
1 46.2
};
\addlegendentry{$L=1$}

\addplot [semithick, color=orange, mark=*]
table {%
0.1 86.1
0.2 86.1
0.3 85.7
0.4 86
0.5 85.5
0.6 83.9
0.7 78.8
0.8 66.8
0.9 52
1 41.3
};
\addlegendentry{$L=2$}

\addplot [semithick, color=red, mark=*]
table {%
0.1 85.8
0.2 85.7
0.3 85.2
0.4 85.1
0.5 84.5
0.6 82.6
0.7 76.2
0.8 64.2
0.9 49
1 39.2
};
\addlegendentry{$L=3$}

\end{axis}
\end{tikzpicture}
    }
    \caption{Hyperparameter evaluation of SGR on the training sets with $k=1$.}\label{fig:eval_gamma}    
\end{figure}
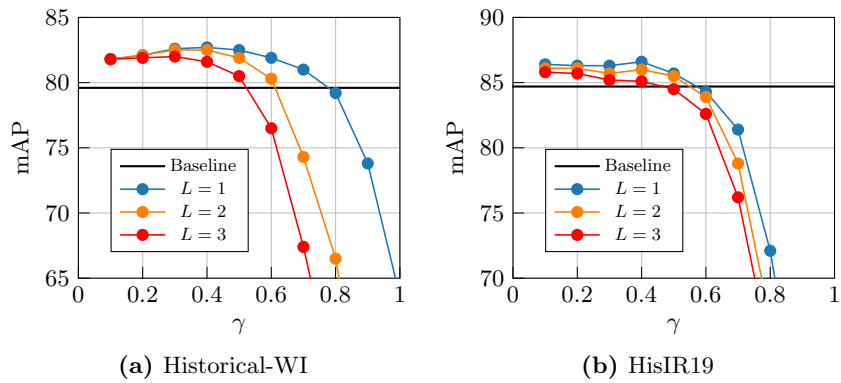

\paragraph{Hyperparameter evaluation} For choosing $L$ and $\gamma$, we perform a grid search on the global descriptors of the training set on both datasets where $\gamma \in [0.1, 1]$, $L \in \{1,2,3\}$. We fix $k=1$ to concentrate on the influence of $\gamma$ and $L$ by prioritizing aggregating correct matches ($n_G = 3$ for the training set of Historical-WI and $n_G \in \{1,3,5\}$ for the training set of HisIR19). The results on both sets are shown in \prettyref{fig:eval_gamma}. Regarding $\gamma$, values up to \num{0.5} improve the baseline performance. Afterward, the \ac{mAP} rapidly drops on both datasets - large values of $\gamma$ also flatten the peaks in the similarity matrix. The influence of the number of layers is smaller when only considering $\gamma \leq \num{0.5}$. However, the best \ac{mAP} is achieved with $L=1$. Therefore, for the evaluation of the test sets, we choose $\gamma = 0.4$ and $L=1$. 

\begin{figure}
    \centering
    \subfloat[Historical-WI]{
        \centering
        \begin{tikzpicture}

\definecolor{darkgray176}{RGB}{176,176,176}
\definecolor{steelblue31119180}{RGB}{31,119,180}

\begin{axis}[
width=0.48\textwidth,
height=0.25\textwidth,
tick align=inside,
tick pos=left,
xmin=0.7, xmax=6.3,
xtick={1,2,3,4,5,6},
xticklabels={1,2,3,4,5,6},
xtick style={color=black},
ymin=70, ymax=85,
ytick style={color=black},
xlabel= $k$,
ylabel=\ac{mAP},
xmajorgrids,
ymajorgrids,
]

\addplot [semithick, color=steelblue31119180, mark=*]
table {%
1 77.6
2 80.6
3 79.2
4 75.8
5 74.7
6 72.8
};

\addplot [mark=none, thick, black]
table {%
0 73.4
7 73.4
};
\end{axis}

\end{tikzpicture}
        \begin{tikzpicture}

\definecolor{darkgray176}{RGB}{176,176,176}
\definecolor{steelblue31119180}{RGB}{31,119,180}

\begin{axis}[
width=0.48\textwidth,
height=0.25\textwidth,
tick align=inside,
tick pos=left,
xmin=0.7, xmax=6.3,
xtick={1,2,3,4,5,6},
xticklabels={1,2,3,4,5,6},
xtick style={color=black},
ytick={85, 90, 95},
yticklabels={85, 90, 95},
ymin=85, ymax=95,
ytick style={color=black},
xlabel= $k$,
ylabel=Top-1,
xmajorgrids,
ymajorgrids,
]

\addplot [mark=none, thick, black]
table {%
0 88.6
7 88.6
};

\addplot [semithick, color=steelblue31119180, mark=*]
table {%
1 88.9
2 91.1
3 92.4
4 92.8
5 92.6
6 91.7
};

\end{axis}

\end{tikzpicture}
    }
    
    \subfloat[HisIR19]{
        \centering
        \begin{tikzpicture}

\definecolor{darkgray176}{RGB}{176,176,176}
\definecolor{steelblue31119180}{RGB}{31,119,180}

\begin{axis}[
legend style={nodes={scale=0.75, transform shape}},
width=0.48\textwidth,
height=0.25\textwidth,
tick align=inside,
xmin=0.7, xmax=6.3,
tick pos=left,
xtick={1,2,3,4,5,6},
xticklabels={1,2,3,4,5,6},
ymin=70, ymax=100,
ytick style={color=black},
xlabel= $k$,
ylabel=\ac{mAP},
xmajorgrids,
ymajorgrids,
legend style={at={(0.1,0.1)},anchor=south west}
]

\addplot [mark=none, thick, black]
table {%
0 91.6
7 91.6
};
\addplot [semithick, color=steelblue31119180, mark=*]
table {%
1 92.5
2 93.2
3 88.2
4 81.3
5 78.9 
6 75.7
};

\end{axis}
\end{tikzpicture}
        \begin{tikzpicture}

\definecolor{darkgray176}{RGB}{176,176,176}
\definecolor{steelblue31119180}{RGB}{31,119,180}

\begin{axis}[
legend style={nodes={scale=0.75, transform shape}},
width=0.48\textwidth,
height=0.25\textwidth,
tick align=inside,
tick pos=left,
xmin=0.7, xmax=6.3,
xtick={1,2,3,4,5,6},
xticklabels={1,2,3,4,5,6},
xtick style={color=black},
ytick={90, 95, 100},
yticklabels={90, 95, 100},
ymin=90, ymax=100,
ytick style={color=black},
xlabel= $k$,
ylabel=Top-1,
xmajorgrids,
ymajorgrids,
legend style={at={(0.1,0.1)},anchor=south west}
]

\addplot [mark=none, thick, black]
table {%
0 96
7 96
};

\addplot [semithick, color=steelblue31119180, mark=*]
table {%
1 96.1
2 96.7
3 97.1
4 97.3
5 96.5
6 95.3
};

\end{axis}

\end{tikzpicture}
    }
    
    \caption{Reranking results of SGR for both datasets. Horizontal lines mark the baseline performance of NetRVLAD. ($\gamma = 0.4$, $L=1$)}\label{fig:eval_reranking}    
\end{figure}
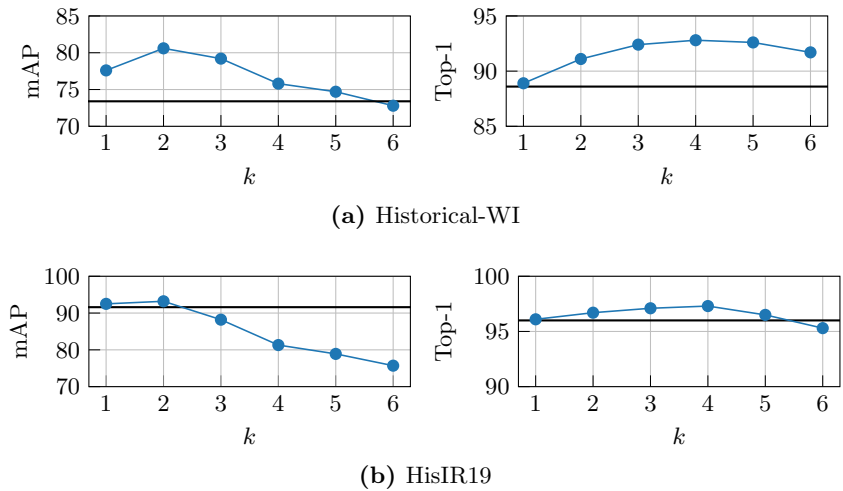

\paragraph{Reranking results}
Finally, we report our results for different values of $k$ on the test set as illustrated in \prettyref{fig:eval_reranking}. The gallery sizes are $n_G = 5$ for Historical-WI and $n_G \in \{1,3,5\}$ for HisIR19. \ac{SGR} boosts \ac{mAP} and Top-1 accuracy, in particular the \ac{mAP} when choosing small values for $k$. For the two datasets with different gallery sizes, the best \ac{mAP} is obtained for $k=2$, for which we achieve $\num{80.6}\%$ and $\num{93.2}\%$ on Historical-WI resp. HisIR19. Afterward, the \ac{mAP} drops on the HisIR19 dataset - we think this is mainly due to the large number of authors contributing only a single document which may be reranked when considering too many neighbors. Interestingly, the Top-1 accuracy even increases for larger values peaking for both datasets at $k=4$ with $\num{92.8}\%$ and $\num{97.3}\%$. 

\subsection{Comparison to State-of-the-art}

We compare our approach concerning two aspects: the performance of the baseline (NetRVLAD) and the reranked descriptors (\ac{SGR}). Our baseline is combined with the graph reranking method in \cite{graphrerank} as well as the \ac{kRNN}-\ac{QE} proposed in \cite{rasoulzadeh}, which is mainly designed for writer retrieval. For both methods \cite{rasoulzadeh} and \cite{graphrerank}, we perform a grid search and report the results of the best hyperparameters to ensure a fair comparison. 

Our feature extraction is similar to Christlein et al. \cite{unsupervised_icdar17} and Chammas et al. \cite{chammas} in terms of preprocessing and training. In contrast, the method proposed in \cite{bVLAD} relies on handcrafted features encoded by multiple VLAD codebooks.

\begin{table}[t]
    \renewcommand{\arraystretch}{1.3}
    \centering
    \caption{Comparison of state-of-the-art methods on Historical-WI. (*) denotes our implementation of the reranking algorithm, (+) reranking applied on the baseline method.}\label{tab:icdar17}
   \begin{tabular}{p{8cm}cc}\hline
        ~ & \ac{mAP} & Top-1 \\ \hline
        CNN+mVLAD \cite{unsupervised_icdar17} & \num{74.8} & \num{88.6} \\
        Pathlet+SIFT+bVLAD \cite{bVLAD} & \num{77.1} & \num{90.1} \\ 
        CNN+mVLAD+ESVM \cite{unsupervised_icdar17} & \num{76.2} & \num{88.9} \\
        \ \ + Pair/Triple SVM \cite{reranking_jordan} & \num{78.2} & \num{89.4} \\
        NetRVLAD (ours) & \num{73.4} & \num{88.5} \\ 
        \ \ + \ac{kRNN}-\ac{QE}* $_{k=3}$ \cite{rasoulzadeh} & 77.1 & 86.8 \\ 
        \ \ + Graph reranking* $_{k_1 = 4, \ k_2 = 2, \ L=3}$ \cite{graphrerank} & 77.6 & 87.4 \\
        \ \ + \ac{SGR} $_{k=2}$ (ours) & \bfseries 80.6 & \bfseries 91.1 \\ \hline
        \end{tabular}
\end{table}

For the Historical-WI dataset, NetRVLAD achieves a \ac{mAP} of $73.4\%$ and, according to \prettyref{tab:icdar17}, our global descriptors are less effective compared to the work of \cite{unsupervised_icdar17}. Regarding reranking, \ac{SGR} outperforms the reranking methods proposed by Jordan et al. \cite{reranking_jordan}, who use a stronger baseline with the mVLAD approach of \cite{unsupervised_icdar17}. Additionally, \ac{SGR} performs better than the graph reranking approach \cite{graphrerank} our method is based on. When using \ac{SGR}, our approach sets a new State-of-the-art performance with a \ac{mAP} of 80.6\% and a Top-1 accuracy of \num{91.1}\%. Compared to the other reranking methods, \ac{SGR} is the only method that improves the Top-1 accuracy.

\begin{table}
    \renewcommand{\arraystretch}{1.3}
    \centering
    \caption{Comparison of state-of-the-art methods on HisIR19. (*) denotes our implementation of the reranking algorithm, (+) reranking applied on the baseline method.}\label{tab:icdar19}
    \begin{tabular}{p{8cm}cc}\hline
        ~ & \ac{mAP} & Top-1 \\ \hline
        CNN+mVLAD \cite{chammas} & \num{91.2} & \num{97.0} \\
        Pathlet+SIFT+bVLAD \cite{bVLAD} & \num{92.5} & \bfseries 97.4 \\
        NetRVLAD (ours) & \num{91.6} & \num{96.1} \\ 
        \ \ + \ac{kRNN}-\ac{QE}* $_{k=4}$ \cite{rasoulzadeh} & 92.6 & 95.2 \\ 
        \ \ + Graph reranking* $_{k_1 = 4, \ k_2 = 2, \ L=2}$ \cite{graphrerank} & 93.0 & 95.7 \\
        \ \ + \ac{SGR} $_{k=2}$ (ours) & \bfseries 93.2 & \num{96.7} \\ \hline
        \end{tabular}
\end{table}

Regarding the performance on the HisIR19 dataset shown in \prettyref{tab:icdar19}, NetRVLAD achieves a \ac{mAP} of 91.6\% and therefore slightly beats the traditional mVLAD method in \cite{chammas}. \ac{SGR} is better than the reranking methods proposed in \cite{graphrerank} and \cite{rasoulzadeh} with a \ac{mAP} of 93.2\%, a new State-of-the-art performance. However, even with reranking, the Top-1 accuracy of NetRVLAD+\ac{SGR} trails the VLAD methods in \cite{chammas,bVLAD}. The improvements of \ac{SGR} are smaller than on the Historical-WI dataset given that the baseline performance is already quite strong with over 90\%, increasing the difficulty of the reranking process. 

\paragraph{ICDAR2013} Finally, to show the versatility of our unsupervised method, we report the performance on the ICDAR2013 dataset \cite{icdar13}, a modern dataset with 250/1000 pages including two English and two Greek texts per writer with only four lines of text each. Although we are limited to less data compared to historical datasets with a large amount of text included in a page, our approach achieves a notable performance (86.1\% \ac{mAP}), in particular Top-1 accuracy (98.5\%), where it outperforms the supervised approach \cite{rasoulzadeh} as shown in \prettyref{tab:icdar13}.

\begin{table}
    \renewcommand{\arraystretch}{1.3}
    \centering
    \caption{Comparison of state-of-the-art methods on ICDAR2013.}\label{tab:icdar13}
    \begin{tabular}{p{8cm}cc}\hline
        ~ & \ac{mAP} & Top-1 \\ \hline
         Zernike+mVLAD \cite{zernike} & 88.0 & \bfseries 99.4 \\
         NetVLAD+\ac{kRNN}-\ac{QE} (supervised) \cite{rasoulzadeh} & \bfseries 97.4 & 97.4 \\ 
         NetRVLAD+\ac{SGR} $_{k=1}$ (ours) & \ 86.1 & \num{98.5} \\ \hline
        \end{tabular}
\end{table}

\section{Conclusion}\label{sec:conclusion}

This paper introduced an unsupervised approach for writer retrieval. We proposed NetRVLAD to directly train the encoding space with $32 \times 32$ patches on labels obtained by clustering their SIFT descriptors. In our experiments, we showed that NetRVLAD outperforms the traditional implementation while also being relatively robust to the codebook's size and backbone architecture. Furthermore, our graph reranking method \ac{SGR} was used to boost the retrieval performance. \ac{SGR} outperformed the original graph reranking and reranking methods recently applied in the domain of writer retrieval. Additionally, we beat the State-of-the-art with our reranking scheme and showed the performance on a modern dataset.

Regarding future work, we think our approach is mainly limited due to the cluster labels used for training. We could overcome this by unlocking the potential of self-supervised methods and train the encoding space without any labels. Other approaches could include learnable poolings, e.g., instead of sum pooling to calculate the global page descriptors, a neural network invariant to permutation could be trained on the patch embeddings to learn a powerful aggregation. Finally, investigating learning-based reranking methods \cite{attention_reranking,reranking_transformers} are a considerable choice for further improving retrieval performance.


\subsubsection*{Acknowledgements} The project has been funded by the Austrian security research programme KIRAS
of the Federal Ministry of Finance (BMF) under the Grant Agreement 879687.

%

\bibliographystyle{splncs04}
\bibliography{bibliography}

\end{document}